\documentclass[11pt,letterpaper]{article}
\usepackage{naaclhlt2015}

\usepackage{times}
\usepackage{latexsym}
\usepackage{amsmath}
\usepackage{multirow}
\usepackage{url}
\usepackage{amsfonts}
\usepackage{amssymb}
\usepackage{graphicx}
\usepackage{mathrsfs}
\usepackage{tablefootnote}
\usepackage{subcaption}

\DeclareMathOperator*{\argmax}{arg\,max}

\title{Unsupervised Dependency Parsing: Let's Use Supervised Parsers}

\author{Phong Le \and Willem Zuidema\\
Institute for Logic, Language, and Computation \\
University of Amsterdam, the Netherlands\\
{\tt \{p.le,zuidema\}@uva.nl } \\ }

\begin{document}

\maketitle

\begin{abstract}

  We present a self-training approach to unsupervised dependency parsing that
  reuses existing supervised and unsupervised parsing algorithms. Our
  approach, called `iterated reranking' (IR), starts
  with dependency trees generated by an unsupervised parser, and
  iteratively improves these trees using the richer probability models
  used in supervised parsing that are in turn trained on these trees. 
  Our system achieves 1.8\%
  accuracy higher than the state-of-the-part parser of Spitkovsky et
  al. (2013) on the WSJ corpus.

\end{abstract}

%%%%%%%%%%%%%%%%%%%%%%%%%%%%%%%%%%%%%%%%%%%%%%%%%%%%%%%%%%%%%%%%%%%%%%%%%%%%%%%%%%
\section{Introduction}

Unsupervised dependency parsing and its supervised counterpart have
many characteristics in common: they take as input raw sentences,
produce dependency structures as output, and often use the same
evaluation metric (DDA, or UAS, the percentage of tokens for which the system
predicts the correct head). Unsurprisingly, there has been much more
research on supervised parsing -- producing a wealth of models,
datasets and training techniques -- than on unsupervised parsing,
which is more difficult, much less accurate and generally uses very
simple probability models.  Surprisingly, however, there have been no
reported attempts to reuse supervised approaches to tackle the
unsupervised parsing problem (an idea briefly mentioned in
\newcite{spitkovsky2010viterbiem}).

There are, nevertheless, two aspects of supervised parsers that we would 
like to exploit in an unsupervised setting. First, we can increase the model
expressiveness in order to capture more linguistic regularities. 
Many recent supervised parsers use third-order (or higher order) features
\cite{koo2010efficient,martins2013turning,le2014the} to 
reach state-of-the-art (SOTA) performance. 
In contrast, existing models for unsupervised parsing 
limit themselves to using simple features (e.g., 
conditioning on heads and valency variables) 
in order to reduce the computational cost, to identify 
consistent patterns in data \cite[page 23]{naseem2014linguistically}, 
and to avoid overfitting \cite{blunsom2010unsupervised}. 
Although this makes learning easier and more efficient, 
the disadvantage is that many useful linguistic regularities are
missed: an upper bound on the performance of such simple models -- estimated by using annotated data -- 
is 76.3\% on the WSJ corpus 
\cite{DBLP:conf/emnlp/SpitkovskyAJ13}, compared to over 93\% 
actual performance of the SOTA supervised parsers.

Second, we would like to make use of information available from lexical semantics, as in 
\newcite{bansal2014tailoring}, \newcite{le2014the}, and \newcite{chen2014}.
Lexical semantics is a source for handling rare 
words and syntactic ambiguities. For instance, if a parser 
can identify that ``he'' is a dependent of ``walks'' in 
the sentence ``He walks'', then, even if ``she'' and ``runs'' 
do not appear in the training data, the parser may still be able 
to recognize that ``she'' should be a dependent of ``runs'' 
in the sentence ``she runs''. Similarly, a parser can make use of the fact that 
``sauce'' and ``John'' have very different meanings to decide  
that they have different heads 
in the two phrases ``ate spaghetti with sauce'' and 
``ate spaghetti with John''. 

However, applying existing supervised parsing techniques to the task of unsupervised parsing is, unfortunately, not trivial. 
The reason is that those parsers are optimally designed for being trained on 
manually annotated data. 
If we use existing unsupervised training methods (like EM), learning could be easily 
misled by a large amount of ambiguity naturally embedded in unannotated 
training data.
Moreover, the computational cost could rapidly increase 
if the training algorithm is not designed properly.
To overcome these difficulties
we propose a framework, iterated reranking (IR), 
where existing supervised parsers are trained 
without the need of manually annotated data, starting with dependency trees provided by an existing unsupervised parser as initialiser.
Using this framework, we can employ the work of \newcite{le2014the}
to build a new system that outperforms the SOTA
unsupervised parser of \newcite{DBLP:conf/emnlp/SpitkovskyAJ13}
on the WSJ corpus. 

The contribution of this paper is twofold. First, we show 
the benefit of using lexical semantics for the unsupervised parsing
task. Second, our work is a bridge connecting the two research 
areas± unsupervised parsing and its supervised counterpart. 
Before going to the next section, in order to avoid confusion introduced by names, 
it is worth noting that we use \textit{un-trained} existing supervised parsers
which will be trained on \textit{automatically annotated} treebanks.

%%%%%%%%%%%%%%%%%%%%%%%%%%%%%%%%%%%%%%%%%%%%%%%%%%%%%%%%%%%%%%%%%%%%%%%%%%%%%%%%%%
\section{Related Work}
\label{section related work}

%-------------------------------------------------------------
\subsection{Unsupervised Dependency Parsing}
\label{section related work - udp}

The first breakthrough was set by \newcite{DBLP:conf/acl/KleinM04}
with their dependency model with valence (DMV), the first 
model to outperform the right-branching baseline on the 
DDA metric: 43.2\% vs 33.6\% on sentences up to length 10 
in the WSJ corpus. Nine years later, \newcite{DBLP:conf/emnlp/SpitkovskyAJ13} 
achieved much higher DDAs: 72.0\% 
on sentences up to length 10, and 64.4\% on all sentences 
in section 23. During this period,  
many approaches have been proposed to attempt the challenge.

\newcite{naseem2011using}, \newcite{tu2012unambiguity}, 
\newcite{spitkovsky2012wabisabi}, \newcite{DBLP:conf/emnlp/SpitkovskyAJ13}, 
and \newcite{DBLP:conf/acl/MarecekS13}
employ extensions of the DMV but with different learning strategies. 
\newcite{naseem2011using} use semantic cues, which are event 
annotations from an out-of-domain annotated corpus, in their model 
during training. 
Relying on the fact that natural language grammars must be unambiguous 
in the sense that a sentence should have very few correct parses, 
\newcite{tu2012unambiguity} incorporate unambiguity regularisation 
to posterior probabilities. \newcite{spitkovsky2012wabisabi} bootstrap 
the learning by slicing up all input sentences at punctuation. 
\newcite{DBLP:conf/emnlp/SpitkovskyAJ13} propose a complete deterministic 
learning framework for breaking out of local optima using count transforms 
and model recombination. \newcite{DBLP:conf/acl/MarecekS13} make use 
of a large raw text corpus (e.g., Wikipedia) to estimate stop probabilities,
using the reducibility principle. 

Differing from those works, 
\newcite{bisk2012simple} rely on Combinatory Categorial Grammars with a 
small number of hand-crafted general linguistic principles; whereas 
\newcite{blunsom2010unsupervised} use Tree Substitution Grammars 
with a hierarchical non-parametric Pitman-Yor process prior 
biasing the learning to a small grammar. 

%------------------------------------------------------------
\subsection{Reranking}

Our work relies on reranking 
which is a technique widely used in (semi-)supervised parsing. 
Reranking requires two components: a $k$-best parser and a reranker. 
Given a sentence, the parser generates a list of $k$ best
candidates, the reranker then rescores those candidates and picks
the one that has the highest score. 
Reranking was first successfully applied to supervised constituent parsing
\cite{DBLP:conf/icml/Collins00,DBLP:conf/acl/CharniakJ05}. 
It was then employed in the supervised dependency parsing approaches 
of \newcite{sangati2009generative}, 
\newcite{hayashi2013efficient}, and \newcite{le2014the}.

Closest to our work is the work series on semi-supervised constituent 
parsing of McClosky and colleagues, e.g. \newcite{mcclosky2006effective}, 
using self-training.
They use a $k$-best generative parser and a discriminative reranker 
to parse unannotated sentences, then add resulting parses to 
the training treebank and re-train the reranker. 
Different from their work, our work is for unsupervised dependency 
parsing, without manually annotated data, and uses iterated reranking
instead of single reranking. In addition, both two components, $k$-best
parser and reranker, are re-trained after each iteration.

%%%%%%%%%%%%%%%%%%%%%%%%%%%%%%%%%%%%%%%%%%%%%%%%%%%%%%%%%%%%%%%%%%%%%%%%%%%%%%%%%%
\section{The IR Framework}
\label{section MPIR}

Existing training methods for the unsupervised dependency 
task, such as \newcite{blunsom2010unsupervised},
\newcite{gillenwater2011posterior}, and \newcite{tu2012unambiguity},
are hypothesis-oriented search with the EM algorithm 
or its variants: training is to move 
from a point which represents a model hypothesis to another point.
This approach is feasible for optimising models 
using simple features since existing dynamic programming 
algorithms can compute expectations, 
which are sums over all possible parses, 
or to find the best parse in the whole parse space with low complexities. 
However, the complexity increases rapidly if rich, complex 
features are used. 
One way to reduce the computational cost is to use approximation 
methods like sampling as in \newcite{blunsom2010unsupervised}. 

\subsection{Treebank-oriented Greedy Search}

Believing that the difficulty of using EM is from the fact that 
treebanks are `hidden', leading to the need of computing sum (or max) overall 
possible treebanks,  
we propose a greedy local search scheme based on another  
training philosophy: treebank-oriented search.
The key idea is to explicitly search for concrete treebanks which 
are used to train parsing models. 
This scheme thus allows supervised parsers to
be trained in an unsupervised parsing setting 
since there is a (automatically annotated) 
treebank at any time. 

Given $\mathcal{S}$ a set of raw sentences, the search 
space consists of all possible treebanks
$\mathcal{D} = \{ d(s) | s \in \mathcal{S} \}$ 
where $d(s)$ is a dependency tree of sentence $s$.
The target of search is the optimal treebank $\mathcal{D}^*$ that is 
as good as human annotations. Greedy search with this 
philosophy is as follows: starting at an initial point 
$\mathcal{D}_1$, we pick up a point $\mathcal{D}_2$ among its 
neighbours $\mathbf{N}(\mathcal{D}_1)$ such that 
\begin{equation}
    \mathcal{D}_2 = \argmax_{\mathcal{D} \in \mathbf{N}(\mathcal{D}_1) } 
                        f_{\mathcal{D}_1} (\mathcal{D})
\end{equation}
where $f_{\mathcal{D}_1} (\mathcal{D})$ is an objective function 
measuring the goodness of $\mathcal{D}$ (which may or may not be conditioned 
on $\mathcal{D}_1$). We then continue this search until some stop 
criterion is satisfied. The crucial factor here is to define 
$\mathbf{N}(\mathcal{D}_i)$ and $f_{\mathcal{D}_i} (\mathcal{D})$. 
Below are two special cases of this scheme.

\paragraph{Semi-supervised parsing using reranking} 
\cite{mcclosky2006effective}. This reranking 
is indeed one-step greedy local search. In this scenario, 
$\mathbf{N}(\mathcal{D}_1)$ is the Cartesian product of $k$-best 
lists generated by a $k$-best parser, and 
$f_{\mathcal{D}_i} (\mathcal{D})$ is a reranker.

\paragraph{Unsupervised parsing with hard-EM}
\cite{spitkovsky2010viterbiem} 
In hard-EM, the target is to maximise 
the following objective function with respect to a parameter 
set $\Theta$
\begin{equation}
    \label{equation hardEM obj}
    L(\mathcal{S} | \Theta) = \sum_{s \in \mathcal{S}} \max_{d \in Dep(s)} 
            \log P_\Theta\big(d\big)
\end{equation}
where $Dep(s)$ is the set of all possible dependency structures of $s$. 
The two EM steps are thus
\begin{itemize}
\item Step 1: 
$\mathcal{D}_{i+1} = \argmax_{\mathcal{D}} P_{\Theta_i}(\mathcal{D})$
\item Step 2: 
$\Theta_{i+1} = \argmax_{\Theta} P_{\Theta}(\mathcal{D}_{i+1})$
\end{itemize}
In this case, $\mathbf{N}(\mathcal{D}_i)$ is the whole treebank space and 
$f_{\mathcal{D}_i}(\mathcal{D}) = P_{\Theta_{i}}(\mathcal{D}) = P_{\argmax_{\Theta} P_{\Theta}(\mathcal{D}_i)}(\mathcal{D})$.

\subsection{Iterated Reranking}

%\begin{figure*}
%\centering
%\includegraphics[width=0.75\textwidth]{IterReranking.png}
%\caption{The iterated reranking. The squares denote \textit{training}, the 
%diamonds denote \textit{parsing}. The pentagons denote \textit{reranking}. }
%\label{figure iterated reranking}
%\end{figure*}

We instantiate the greedy search scheme by iterated reranking
%(see Figure~\ref{figure iterated reranking}) 
which requires two components: a $k$-best parser $P$, and a reranker $R$. 
Firstly, $\mathcal{D}_1$ is used to train these two components, 
resulting in $P_1$ and $R_1$. The parser $P_1$ then generates 
a set of lists of $k$ candidates $_k\mathcal{D}_{1}$ (whose Cartesian 
product results in $\mathbf{N}(\mathcal{D}_1)$) for the set of 
training sentences $\mathcal{S}$. The best candidates, according 
to reranker $R_1$, are collected to form $\mathcal{D}_2$ for the 
next iteration. This process is halted when a pre-defined stop 
criterion is met.\footnote{
It is worth noting that, although $\mathbf{N}(\mathcal{D}_i)$ 
has the size $O(k^n)$ where $n$ is the number of sentences, reranking 
only needs to process $O(k\times n)$ parses if these sentences 
are assumed to be independent.}

It is certain that we can, as in the work of 
\newcite{spitkovsky2010viterbiem} and many bootstrapping approaches, 
employ only parser $P$. Reranking, however, brings us two benefits. 
First, it allows us to employ very expressive models like the $\infty$-order 
generative model proposed by \newcite{le2014the}. Second, 
it embodies a similar idea to co-training
\cite{DBLP:conf/colt/BlumM98}: $P$ and $R$ play roles as two 
views of the data. 
%As \newcite{sangati2009generative}
%suggest, if these two views are orthogonal, resulting 
%parses are more correct. 

\subsection{Multi-phase Iterated Reranking}

%\begin{figure*}
%\centering
%\includegraphics[width=1.\textwidth]{MultiPhaseIR.png}
%\caption{The multi-phase iterated reranking. Each phase is an iterated reranking shown in 
%Figure~\ref{figure iterated reranking}. We use the resulting parser in the previous phase to 
%generate the starting point for the iterated reranking in the next phase. }
%\label{figure multi-phase IR}
%\end{figure*}

Training in machine learning often uses  
\textit{starting big} which is to use up all training data at 
the same time. However, \newcite{elman1993learning} suggests 
that in some cases, learning should start by training simple 
models on small data and then gradually increase the model
complexity and add more difficult data. This is called 
\textit{starting small}. 

In unsupervised dependency parsing, starting small is 
intuitive. For instance, given a set of long sentences, learning 
the fact that the head of a sentence is its main verb is 
difficult because a long sentence always contains many syntactic 
categories. It would be much easier if we start with only length-one 
sentences, e.g ``Look!'', since there is only one choice which is 
usually a verb. This training scheme was successfully applied 
by \newcite{SpitkovskyEtAl10} under the name: Baby Step.

We adopt starting small to construct the multi-phase iterated reranking 
(MPIR) framework.
%(see Figure~\ref{figure multi-phase IR})
In phase 0, a parser $M$ with a simple model  is trained on 
a set of short sentences $\mathcal{S}^{(0)}$ as in traditional 
approaches. This parser is used to parse a larger set of sentences 
$\mathcal{S}^{(1)} \supseteq \mathcal{S}^{(0)}$, resulting in 
$\mathcal{D}^{(1)}_1$. $\mathcal{D}^{(1)}_1$ is then used as 
the starting point for the iterated reranking in phase 1. 
We continue this process until phase $N$ finishes, with 
$\mathcal{S}^{(i)} \supseteq \mathcal{S}^{(i-1)}$ ($i=1..N$). 
In general, we use the resulting reranker in the previous phase 
to generate the starting point for the iterated reranking in the 
current phase. 

%%%%%%%%%%%%%%%%%%%%%%%%%%%%%%%%%%%%%%%%%%%%%%%%%%%%%%%%%%%%%%%%%%%%%%%%%%%%%
\section{\newcite{le2014the}'s Reranker}
\label{section reranker}

\newcite{le2014the}'s reranker is an exception among supervised 
parsers because it employs an extremely \textit{expressive} model
whose features are $\infty$-order\footnote{In fact, the order is finite but unbound.}. 
To overcome the problem of 
sparsity, they introduced the inside-outside recursive
neural network (IORNN) architecture that can estimate 
tree-generating models including those proposed by 
\newcite{eisner1996three} and \newcite{collins2003head}. 

\subsection{The $\infty$-order Generative Model}

\newcite{le2014the}'s reranker employs the generative model proposed by 
\newcite{eisner1996three}. Intuitively, this model is top-down: starting 
with ROOT, we generate its left dependents and its right dependents. 
We then generate dependents for each ROOT's dependent. The generative 
process recursively continues until there is no dependent to generate. 
Formally, this model is described by the following formula
{\small
\begin{align} \label{equ gen}
    \footnotesize
    P(d(H)) &= \prod_{l=1}^L P\left( H^L_l | \mathcal{C}(H^L_l) \right)  
                    P\left( d(H^L_l) \right) \times \notag \\
            &\prod_{r=1}^R P\left( H^R_r | \mathcal{C}(H^R_r) \right)  
                    P\left( d(H^R_r) \right)
\end{align}
}
where $H$ is the current head, $d(N)$ is the fragment of the dependency 
parse rooted at $N$, and $\mathcal{C}(N)$ is the context to generate $N$.  
$H^L, H^R$ are respectively $H$'s left dependents and right dependents, 
plus $EOC$ (End-Of-Children), a special token to inform that there are 
no more dependents to generate. Thus, $P(d(ROOT))$ is the probability 
of generating the entire dependency structure $d$.

Le and Zuidema's $\infty$-order generative model is defined as 
Eisner's model in which the context $\mathcal{C}^{\infty}(D)$ 
to generate $D$ contains \textit{all} of $D$'s generated siblings, 
its ancestors and their siblings.
%(the fragment enclosed 
%in the red dashed contour in Figure~\ref{figure orders}).
Because of very large fragments that contexts are allowed to hold, 
traditional count-based methods are impractical (even if we use smart 
smoothing techniques). They thus introduced the
IORNN architecture to estimate the model.
%\begin{figure*}
%\centering
%\includegraphics[width=0.8\textwidth]{dev.png}
%\caption{Example of different orders of context of ``diversified''. 
%The blue doted shape corresponds to the third-order outward context, 
%while the red dashed shape corresponds to the $\infty$-order left-to-right context.
%The green dot-dashed shape corresponds to the context to compute the 
%outer representation. }
%\label{figure orders}
%\end{figure*}

\subsection{Estimation with the IORNN}

\begin{figure}
    \centering
    \includegraphics[scale=0.35]{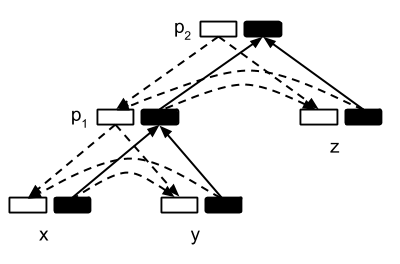}
    \caption{Inside-Outside Recursive Neural Network (IORNN).
            Black/white rectangles correspond to inner/outer representations.}
    \label{figure iornn}
\end{figure}

An IORNN (Figure~\ref{figure iornn}) is a recursive neural network 
whose topology is a tree. What make this network different from 
traditional RNNs \cite{socher_learning_2010} is that each tree node 
$u$ caries two vectors: $\mathbf{i}_u$ - the inner 
representation, represents the content of the phrase covered 
by the node, and $\mathbf{o}_u$ - the outer representation, 
represents the context around that phrase. In addition, information 
in an IORNN is allowed to flow not only bottom-up as in RNNs, but 
also top-down. That makes IORNNs a natural tool for estimating 
top-down tree-generating models. 

Applying the IORNN architecture to dependency parsing is straightforward, 
along the generative story of the $\infty$-order generative model. 
First of all, the ``inside'' part of this IORNN is simpler than 
what is depicted in Figure~\ref{figure iornn}: the inner representation 
of a phrase is assumed to be the inner representation of its head. 
This approximation is plausible since the meaning of a phrase is often 
dominated by the meaning of its head. The inner representation at 
each node, in turn, is a function of a vector 
representation for the word (in our case, the word vectors are initially
borrowed from \newcite{collobert_natural_2011}), the POS-tag and
capitalisation feature.

Without loss of generality and ignoring directions for simplicity, 
they assume that the model is generating 
dependent $u$ for node $h$ conditioning on context $\mathcal{C}^{\infty}(u)$ 
which contains all of $u$'s ancestors (including $h$) and theirs 
siblings, and all of previously generated $u$'s sisters. 
Now there are two types of contexts: \textit{full} contexts of heads (e.g., $h$)
whose dependents are being generated, and contexts to generate nodes
(e.g., $\mathcal{C}^\infty(u)$). 
Contexts of the first type are clearly represented by outer 
representations. Contexts of the other type are represented by 
\textit{partial outer representations}, denoted by $\bar{\mathbf{o}}_u$.
Because the context to generate a node can be constructed recursively by 
combining the full context of its head and its previously generated sisters, 
they can compute $\bar{\mathbf{o}}_u$ as a function of $\mathbf{o}_h$ and 
the inner representations of its previously generated sisters. 
On the top of $\bar{\mathbf{o}}_u$, they put a softmax layer to estimate 
the probability $P(x|\mathcal{C}^\infty(u))$. 

Training this IORNN is to minimise the cross entropy over all dependents. 
This objective function is indeed the negative log likelihood $P(\mathcal{D})$ 
of training treebank $\mathcal{D}$.

\subsection{The Reranker}
\label{subsection reranker}
Le and Zuidema's (generative) reranker is given by 
\begin{equation*}
    d^* = \argmax_{d \in _k Dep(s)} P(d) 
\end{equation*}
where $P$ (Equation~\ref{equ gen}) is computed by the $\infty$-order generative model 
which is estimated by an IORNN; 
and $_k Dep(s)$ is a $k$-best list.

%%%%%%%%%%%%%%%%%%%%%%%%%%%%%%%%%%%%%%%%%%%%%%%%%%%%%%%%%%%%%%%%%%%%%%%%%%%%%%%%%%
\section{Complete System}
\label{section system}

Our system is based on the multi-phase IR.
In general, any third-party parser for unsupervised 
dependency parsing can be used in phase 0, and any third-party 
parser that can generate $k$-best lists can be used in the other 
phases. In our experiments, for phase 0, we choose the parser 
using an extension of the DMV model with stop-probability estimates
computed on a large corpus proposed by \newcite{DBLP:conf/acl/MarecekS13}. 
This system has a moderate performance\footnote{
\label{footnote mz}\newcite{DBLP:conf/acl/MarecekS13}
did not report any experimental result on the WSJ corpus. We use 
their source code at \url{http://ufal.mff.cuni.cz/udp}
with the setting presented in Section~\ref{subsection setting}. 
Because the parser does not provide the option to parse unseen 
sentences, we merge the training sentences (up to length 15) to all 
the test sentences to evaluate its performance. Note that 
this result is close to the DDA (55.4\%) that the authors reported on 
CoNLL 2007 English dataset, which is a portion of the WSJ corpus.}
on the WSJ corpus: 57.1\% vs the SOTA 64.4\% DDA of 
\newcite{DBLP:conf/emnlp/SpitkovskyAJ13}. For the other phases, 
we use the MSTParser\footnote{\url{http://sourceforge.net/projects/mstparser/}}
(with the second-order feature mode) \cite{DBLP:conf/eacl/McDonaldP06}.

Our system uses \newcite{le2014the}'s reranker (Section~\ref{subsection reranker}). 
It is worth noting that, in this case, each phase with iterated 
reranking could be seen as an approximation of hard-EM 
(see Equation~\ref{equation hardEM obj}) where 
the first step is replaced by 
\begin{equation}
    \mathcal{D}_{i+1} = \argmax_{\mathcal{D} \in \mathbf{N}(\mathcal{D}_i)} 
                                    P_{\Theta_i}(\mathcal{D})
    \label{equation approx hard-em}
\end{equation}
In other words, instead of searching over the treebank space, 
the search is limited in a neighbour set $\mathbf{N}(\mathcal{D}_i)$ 
generated by $k$-best parser $P_i$.

\subsection{Tuning Parser $P$}
\label{subsection tuning parser}
Parser $P_i$ trained on $\mathcal{D}_i$ defines neighbour set
$\mathbf{N}(\mathcal{D}_i)$ which is the Cartesian product
of the $k$-best lists in $_k\mathcal{D}_i$. The position and shape 
of $\mathbf{N}(\mathcal{D}_i)$ is thus determined by two factors:  
how well $P_i$ can fit $\mathcal{D}_i$, and $k$. Intuitively, 
the lower the fitness is, the more $\mathbf{N}(\mathcal{D}_i)$
goes far away from $\mathcal{D}_i$; and the larger $k$ is, 
the larger $\mathbf{N}(\mathcal{D}_i)$
is. Moreover, the diversity of $\mathbf{N}(\mathcal{D}_i)$ is inversely 
proportional to the fitness. When the fitness decreases, 
patterns existing in the training treebank become less certain
to the parser, patterns that do not exist in the training treebank 
thus have more chances to appear in $k$-best candidates. This 
leads to high diversity of $\mathbf{N}(\mathcal{D}_i)$.
We blindly set $k=10$ in all of our experiments. 

With the MSTParser, there are two hyper-parameters: 
\texttt{iters}$_\text{MST}$, the number of epochs, and 
\texttt{training-k}$_\text{MST}$, the $k$-best parse set size to 
create constraints during training. \texttt{training-k}$_\text{MST}$ 
is always 1 because constraints from $k$-best parses with 
almost incorrect training parses are useless.

Because \texttt{iters}$_\text{MST}$ controls the fitness of the 
parser to training treebank $\mathcal{D}_i$, it, as pointed 
out above, determines the distance from $\mathbf{N}(\mathcal{D}_i)$
to $\mathcal{D}_i$ and the diversity of the former. Therefore, if we 
want to encourage the local search to explore more distant areas, we 
should set \texttt{iters}$_\text{MST}$ low. In our experiments, 
we test two strategies: (i) MaxEnc, \texttt{iters}$_\text{MST}$ = 1, 
maximal encouragement, and (ii) MinEnc, \texttt{iters}$_\text{MST}$ = 10, 
minimal encouragement.

\subsection{Tuning Reranker $R$}

Tuning the reranker $R$ is to set values for \texttt{dim}$_\text{IORNN}$, 
the dimensions of inner and outer representations, and 
\texttt{iters}$_\text{IORNN}$, the number of epochs to train the IORNN. 
Because the $\infty$-order model is very expressive and feed-forward 
neural networks are universal approximators 
\cite{cybenko1989approximation}, the reranker is capable of perfectly 
remembering all training parses. In order to avoid this, we set 
\texttt{dim}$_\text{IORNN}$ = 50, and set \texttt{iters}$_\text{IORNN}$ = 5 
for \textit{very} early stopping.

\subsection{Tuning multi-phase IR}

Because \newcite{DBLP:conf/acl/MarecekS13}'s parser does not distinguish 
training data from test data, we postulate $\mathcal{S}_0 = \mathcal{S}_1$. 
Our system has $N$ phases such that $\mathcal{S}_0, \mathcal{S}_1$ 
contain all sentences up to length $l_1 = 15$, $\mathcal{S}_i$ ($i=2..N$)
contains all sentences up to length $l_i = l_{i-1} + 1$, 
and $\mathcal{S}_N$ contains all sentences up to length 25.
Phase 1 halts after 100 iterations whereas all the following phases run 
with one iteration. Note that we force the local search in phase 1 to run 
intensively because we hypothesise that most of the important patterns for 
dependency parsing can be found within short sentences.

%%%%%%%%%%%%%%%%%%%%%%%%%%%%%%%%%%%%%%%%%%%%%%%%%%%%%%%%%%%%%%%%%%%%%%%%%%%%%%%%%%
\section{Experiments}
\label{section experiments}

\subsection{Setting}
\label{subsection setting}

We use the Penn Treebank WSJ corpus: sections 02-21  
for training, and section 23 for testing. We then apply the standard
pre-processing\footnote{ 
\url{http://www.cs.famaf.unc.edu.ar/~francolq/en/proyectos/dmvccm}}
for unsupervised dependency parsing task \cite{DBLP:conf/acl/KleinM04}: 
we strip off all empty sub-trees, punctuation, and 
terminals (tagged \# and \$) not pronounced where they appear; we 
then convert the remaining trees to dependencies using Collins's 
head rules \cite{DBLP:journals/coling/Collins03}. Both word forms and 
gold POS tags are used. The directed dependency accuracy (DDA) metric 
is used for evaluation.

The vocabulary is taken as a list of words occurring more than 
two times in the training data. All other words are labelled `UNKNOWN' 
and every digit is replaced by `0'. 
%The whole system is implemented 
%in Torch7 \cite{collobert2011torch7},\footnote{\url{http://www.torch.ch/}}
%which is a powerful Matlab-like environment for machine learning.
We initialise the IORNN with the 50-dim word embeddings from 
\newcite{collobert_natural_2011} \footnote{\url{http://ml.nec-labs.com/senna/}. 
These word embeddings were \textit{unsupervisedly} learnt from Wikipedia.} ,
and train it with the learning rate $0.1$,

%-------------------------------------------------
\subsection{Results}

We compare our system against recent systems 
(Table~\ref{table recent systems} and Section~\ref{section related work - udp}).
Our system with the two encouragement levels, MinEnc and MaxEnc, 
achieves the highest reported DDAs on section 23: 1.8\%  
and 1.2\% higher than \newcite{DBLP:conf/emnlp/SpitkovskyAJ13} 
on all sentences and up to length 10, respectively. 
Our improvements over the system's initialiser \cite{DBLP:conf/acl/MarecekS13}
are 9.1\% and 4.4\%.

\begin{table}
\centering
\begin{tabular}{r|c}
System & DDA (@10) \\ \hline

\newcite{bisk2012simple} & 53.3 (71.5) \\
\newcite{blunsom2010unsupervised} & 55.7 (67.7) \\
\newcite{tu2012unambiguity} & 57.0 (71.4) \\
\newcite{DBLP:conf/acl/MarecekS13} & 57.1 (68.8) \\
\newcite{naseem2011using} & 59.4 (70.2) \\
\newcite{spitkovsky2012wabisabi} & 61.2 (71.4) \\
\newcite{DBLP:conf/emnlp/SpitkovskyAJ13} & 64.4 (72.0) \\

\hline
Our system (MinEnc) & \textbf{66.2} (72.7) \\
Our system (MaxEnc) & 65.8 (\textbf{73.2}) \\

\end{tabular}
\caption{Performance on section 23 of the 
WSJ corpus (all sentences and up to length 10) for recent systems 
and our system. MinEnc and MaxEnc denote \texttt{iters}$_\text{MST}$ = 10 
and \texttt{iters}$_\text{MST}$ = 1 respectively.}
\label{table recent systems}
\end{table}

%-------------------------------------------------
\subsection{Analysis}

In this section, we analyse our system along two aspects. First,
we examine three factors which determine the performance of the whole system: 
encouragement level, lexical semantics, 
and starting point. We then search for what IR (with the MaxEnc option) 
contributes to the overall performance by comparing the quality of the 
treebank resulted in the end of phase 1 against 
the quality of the treebank given by its initialier, 
i.e. \newcite{DBLP:conf/acl/MarecekS13}. 

\subsubsection*{The effect of encouragement level}
\begin{figure}[t!]
\centering
\includegraphics[width=0.45\textwidth]{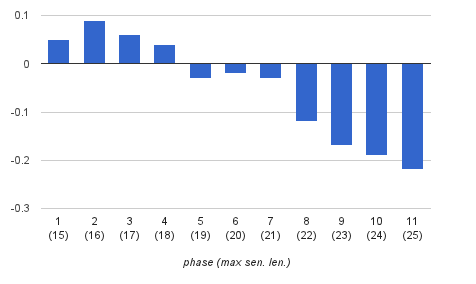}
\caption{$DDA_\textit{MaxEnc} - DDA_\textit{MinEnc}$ of all phases 
on the their training sets (e.g., phase $3$ with $\mathcal{S}^{(3)}$ 
containing all training sentences up to length 17).}
\label{figure phase}
\end{figure}

Figure~\ref{figure phase} shows the differences in DDA between 
using MaxEnc and MinEnc in each phase: we compute 
$DDA_\textit{MaxEnc} - DDA_\textit{MinEnc}$ of each phase on its 
training set 
(e.g., phase $3$ with $\mathcal{S}^{(3)}$ containing all training 
sentences up to length 17). MinEnc outperforms MaxEnc 
within phases 1, 2, 3, and 4. However, from phase 5, the latter 
surpasses the former. It suggests that exploring areas far away 
from the current point with long sentences is risky. The reason 
is that long sentences contain more ambiguities than short ones; 
thus rich diversity, high difference from the current point, 
but small size (i.e., small $k$) could easily lead the learning 
to a wrong path.

The performance of the system with the two encouragement levels 
on section 23 (Table~\ref{table recent systems}) also suggests 
the same. MaxEnc strategy helps the system achieve the highest 
accuracy on short sentences (up to length 10). However, it is less 
helpful than MinEnc when performing on long sentences.

\subsubsection*{The role of lexical semantics}
\begin{figure}[t!]
\centering
\includegraphics[width=0.45\textwidth]{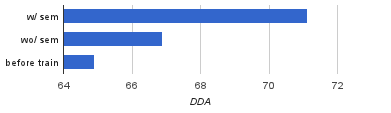}
\caption{DDA of phase 1 (MaxEnc), with and without the word embeddings 
(denoted by w/ sem and wo/ sem, respectively), on training sentences up to 
length 15 (i.e. $\mathcal{S}^{(1)}$). }
\label{figure semantics}
\end{figure}

We examine the role of the lexical semantics, 
which is given by the word embeddings. Figure~\ref{figure semantics} 
shows DDAs on training sentences up to length 15 (i.e. $\mathcal{S}^{(1)}$)
of phase 1 (MaxEnc) with and without the word-embeddings. With the 
word-embeddings, phase 1 achieves 71.11\%. When the word-embeddings 
are not given, i.e. the IORNN uses randomly generated word vectors, 
the accuracy drops 4.2\%. It shows that lexical semantics plays a 
decisive role in the performance of the system. 

However, it is worth noting that, even without that knowledge 
(i.e., with the $\infty$-order generative model alone), 
the DDA of phase 1 is 2\% higher than before being trained
(66.89\% vs 64.9\%). It suggests that phase 1 is capable of discovering 
some useful dependency patterns that are invisible to the parser in 
phase 0. This, we conjecture, is thanks to high-order features captured 
by the IORNN.

\subsubsection*{The importance of the starting point}

\begin{figure}[t!]
\centering
\includegraphics[width=0.5\textwidth]{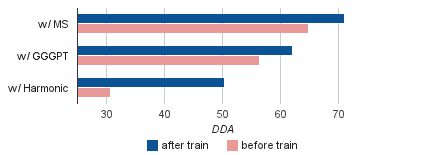}
\caption{DDA of phase 1 (MaxEnc) before and after training 
with three different starting points provided by 
three parsers used in phase 0: MS \cite{DBLP:conf/acl/MarecekS13}, 
GGGPT \cite{gillenwater2011posterior}, and Harmonic \cite{DBLP:conf/acl/KleinM04}.}
\label{figure starting-point}
\end{figure}

Starting point is claimed to be important in local search. 
We examine this by using three different parsers in phase 0: 
(i) MS \cite{DBLP:conf/acl/MarecekS13}, the parser used in the 
previous experiments, (ii) GGGPT
\cite{gillenwater2011posterior}\footnote{\url{code.google.com/p/pr-toolkit}}
employing an extension of the DMV model and posterior 
regularization framework for training, and (iii) Harmonic, 
the harmonic initializer  proposed by \newcite{DBLP:conf/acl/KleinM04}.

Figure~\ref{figure starting-point} shows DDAs of phase 1 (MaxEnc) 
on training sentences up to length 15 with three starting-points 
given by those parsers. Starting point is clearly very important 
to the performance of the iterated reranking: the better the 
starting point is, the higher performance phase 1 has. However, 
a remarkable point here is that the iterated reranking of phase 1 always 
finds out more useful patterns for parsing whatever the starting point is  
in this experiment. It is certainly due to the high order features and   
lexical semantics, which are not exploited in those parsers. 

\subsubsection*{The contribution of Iterated Reranking}
\begin{figure}[t!]
\centering
\begin{subfigure}[b]{0.4\textwidth}
\includegraphics[width=\textwidth]{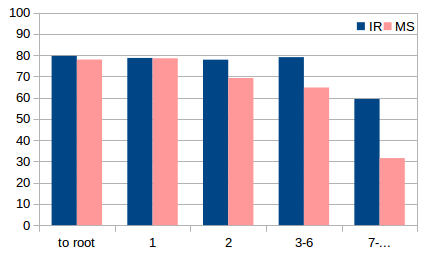}
\end{subfigure}

\begin{subfigure}[b]{0.4\textwidth}
\includegraphics[width=\textwidth]{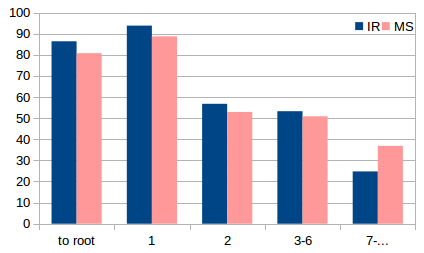}
\end{subfigure}
\caption{Precision (top) and recall (bottom) over binned HEAD distance
of iterated reranking (IR) and its initializer (MS) on the training sentences 
in phase 1 ($\le 15$ words).}
\label{figure prec-recall}
\end{figure}

\begin{figure*}[t!]
\centering
\includegraphics[width=1.\textwidth]{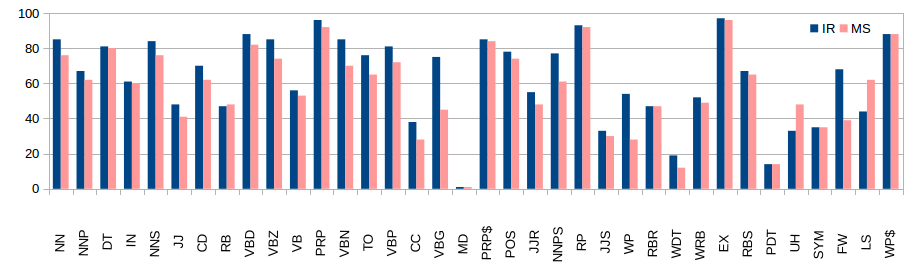}
\caption{Correct-head accuracies over POS-tags (sorted in the descending order 
by frequency) of iterated reranking (IR) 
and its initializer (MS) on the training sentences in phase 1 
($\le 15$ words).}
\label{figure pos-correct}
\end{figure*}

We compare the quality of the treebank resulted in the end of 
phase 1 against the quality of the treebank given by the initialier
\newcite{DBLP:conf/acl/MarecekS13}. 
Figure~\ref{figure prec-recall} shows precision (top) and recall (bottom) 
over binned HEAD distance. IR helps to improve the precision on all distance bins, 
especially on the bins corresponding to long distances ($\ge 3$). 
The recall is also improved, except on the bin corresponding to 
$\ge 7$ (but the F1-score on this bin is increased). We attribute 
this improvement to the $\infty$-order model which uses very large fragments 
as contexts thus be able to capture long dependencies.

Figure~\ref{figure pos-correct} shows the correct-head accuracies over 
POS-tags. IR helps to improve the accuracies over almost all 
POS-tags, particularly nouns (e.g. NN, NNP, NNS), verbs (e.g. VBD, 
VBZ, VBN, VBG) and adjectives (e.g. JJ, JJR). However, as being affected 
by the initializer, IR performs poorly on conjunction (CC) and 
modal auxiliary (MD). For instance, in the treebank given by the initializer, 
almost all modal auxilaries are dependents of their verbs instead of the other way 
around.

\section{Discussion}

Our system is different from the other systems shown in 
Table~\ref{table recent systems} as it uses an extremely 
expressive model, the $\infty$-order generative model, 
in which conditioning contexts are very large fragments. 
Only the work of \newcite{blunsom2010unsupervised}, whose 
resulting grammar rules can contain large tree fragments, 
shares this property. The difference is that their work 
needs a pre-defined prior, namely hierarchical non-parametric 
Pitman-Yor process prior, to avoid large, rare fragments and 
for smoothing. The IORNN of our system, in contrast, does 
that automatically. It learns by itself how to deal with 
distant conditioning nodes, which are often less informative 
than close conditioning nodes on computing 
$P(x|\mathcal{C}^\infty(u))$. In addition, smoothing is 
given free: recursive neural nets are able to map 
`similar' fragments onto close points \cite{socher_learning_2010}
thus an unseen fragment tends to be mapped onto a point close 
to points corresponding to `similar' seen fragments.

Another difference is that our system exploits lexical 
semantics via word embeddings, which were learnt unsupervisedly. 
By initialising the IORNN with these embeddings, 
the use of this knowledge turns out easy and transparent. 
\newcite{DBLP:conf/emnlp/SpitkovskyAJ13} also 
exploit lexical semantics but in a limited way, using
a context-based polysemous unsupervised clustering method 
to tag words. Although their approach can distinguish  
polysemes (e.g., `cool' in `to cool the selling panic' and in 
`it is cool'), it is not able to make use of word meaning 
similarities (e.g., the meaning of `dog' is closer to `animal' than 
to `table'). \newcite{naseem2011using}'s system uses semantic cues 
from an out-of-domain annotated corpus, thus is not fully 
unsupervised.

We have showed that IR with a generative 
reranker is an approximation of hard-EM 
(see Equation~\ref{equation approx hard-em}). 
Our system is thus related to the works of \newcite{DBLP:conf/emnlp/SpitkovskyAJ13} 
and \newcite{tu2012unambiguity}. However, what we have proposed 
is more than that: IR is a 
general framework that we can have more than one option for choosing 
$k$-best parser and reranker. For instance, we can make use of a 
generative $k$-best parser and a discriminative reranker that 
are used for supervised parsing. Our future work is to explore this.

The experimental results reveal that starting point is very important
to the iterated reranking with the $\infty$-order generative model.
On the one hand, that is a disadvantage compared to the other systems, 
which use uninformed or harmonic initialisers. But on the other hand, 
that is an innovation as our approach is capable of making use of 
existing systems. The results shown in Figure~\ref{figure starting-point}
suggest that if phase 0 uses a better parser which uses less expressive model 
and/or less external knowledge than our model, such as the one proposed by
\newcite{DBLP:conf/emnlp/SpitkovskyAJ13}, 
we can expect even a higher performance. The other systems, 
except \newcite{blunsom2010unsupervised}, however, might not benefit from 
using good existing parsers as initializers because their 
models are not significantly more expressive than others \footnote{
In an experiment, we used the \newcite{DBLP:conf/acl/MarecekS13}'s parser 
as an initializer for the \newcite{gillenwater2011posterior}'s parser. As we
expected, the latter was not able to make use of this. }.

%%%%%%%%%%%%%%%%%%%%%%%%%%%%%%%%%%%%%%%%%%%%%%%%%%%%%%%%%%%%%%%%%%%%%%%%%%%%%%%%%%
\section{Conclusion}

We have proposed a new framework, iterated reranking (IR), 
which trains supervised parsers without the need of manually 
annotated data by using a unsupervised parser as an initialiser. 
Our system, employing \newcite{DBLP:conf/acl/MarecekS13}'s unsupervised parser
as the initialiser, the $k$-best MSTParser, and 
\newcite{le2014the}'s reranker, achieved 1.8\% DDA 
higher than the SOTA parser of 
\newcite{DBLP:conf/emnlp/SpitkovskyAJ13} on the WSJ corpus.
Moreover, we also showed that unsupervised parsing benefits 
from lexical semantics through using word-embeddings. 

Our future work is to exploit other existing supervised parsers that 
fit our framework. Besides, taking into account the fast development 
of the word embedding research \cite{mikolov2013distributed,pennington2014glove}, 
we will try different word embeddings. 

\section*{Acknowledgments} 
We thank Remko Scha and three anonymous reviewers for helpful comments. 
Le thanks Milo$\breve{s}$ Stanojevi$\acute{c}$ for helpful discussion. 

\newpage
\bibliographystyle{naaclhlt2015}
\bibliography{ref}

\end{document}